# IFR-Net: Iterative Feature Refinement Network for Compressed Sensing MRI

Yiling Liu, Qiegen Liu, Minghui Zhang*, Qingxin Yang, Shanshan Wang and Dong Liang*, *Senior Member, IEEE*

*Abstract*—To improve the compressive sensing MRI (CS-MRI) approaches in terms of fine structure loss under high acceleration factors, we have proposed an iterative feature refinement model (IFR-CS), equipped with fixed transforms, to restore the meaningful structures and details. Nevertheless, the proposed IFR-CS still has some limitations, such as the selection of hyper-parameters, a lengthy reconstruction time, and the fixed sparsifying transform. To alleviate these issues, we unroll the iterative feature refinement procedures in IFR-CS to a supervised model-driven network, dubbed IFR-Net. Equipped with training data pairs, both regularization parameter and the utmost feature refinement operator in IFR-CS become trainable. Additionally, inspired by the powerful representation capability of convolutional neural network (CNN), CNN-based inversion blocks are explored in the sparsity-promoting denoising module to generalize the sparsity-enforcing operator. Extensive experiments on both simulated and *in vivo* MR datasets have shown that the proposed network possesses a strong capability to capture image details and preserve well the structural information with fast reconstruction speed.

## I. INTRODUCTION

Magnetic resonance imaging (MRI) is a non-invasive and widely used imaging technique that can provide both functional and anatomical information for clinical diagnosis. However, the slow imaging speed may result in patient discomfort and motion artifacts. Therefore, increasing MR imaging speed is an important and worthwhile research goal.

During the past decades, compressed sensing (CS) has become a popular and successful strategy for fast MR imaging reconstruction [1]-[6]. In most early CS-MRI methods, the model consists of two components, namely, the data fidelity term in the *k*-space domain and the regularization term equipped with fixed sparsifying transforms (e.g., finite-difference and wavelet). However, although these classical methods can efficiently solve the problem, losing structures and details can be observed in the reconstructed images, particularly at high undersampling rates. To mitigate this problem, many studies have been conducted, which have mainly focused on two aspects [7]-[20]. One is developing more powerful sparsity transforms or applying nonlocal operations, e.g., discrete cosine transforms [7], [8], patch-based directional wavelets [9], data-driven tight frame [10], dictionary learning [11]-[14], and nonlocal total variation regularization [15]. Although these strategies have improved the reconstruction accuracy, they are time consuming owing to the online training or significant number of nonlocal operations required. Another aspect is to restore the fine structures that are discarded using basic CS-MRI methods [20]. Nevertheless, while bringing back the useful structures, these methods may also introduce noise into the restored image. To alleviate this issue, we proposed an iterative feature refinement method, called IFR-CS, for accurate undersampled MR image reconstruction [21], [22]. IFR-CS extracts only fine structures and details from the residual image, and adds them back to the intermediately denoised image by designing a feature descriptor. Although the desired results can be achieved, IFR-CS and other CS-MRI methods still have some limitations. For example, the iterative approach takes a relatively long time to achieve high-quality image reconstruction. In addition, the selection of the regularization parameter is empirical.

Deep neural networks have recently achieved exciting successes for image classification, segmentation, denoising and accelerated MRI [23]-[40], owing to their strong learning ability from data. Fast online restoration and powerful non-linear mapping ability are the main advantages of deep learning methods. However, the usual data-driven networks rely on the use of large datasets and most medical data are difficult to obtain due to patient privacy. One way to mitigate this problem is training the network with natural images and applying the trained network to restore medical images [40]. Another way to alleviate the problem is recovering medical images with the model-based deep learning networks [41] - [46]. For example, Sun *et al.* proposed a model-based ADMM-Net for CS-MRI [41] by unrolling the procedure of the alternating direction method of multipliers (ADMM) to an iterative network. Specially, this network learns the parameters of the regularization term and transforms in the model through network training, which enables the network to restore the image with fast speed and relieve the parameter selection procedure. Recently, there are more efforts conducted on the model-based networks. Some

This work was in part supported by National Natural Science Foundation of China (61365013, 61503176, 61601450), Basic Research Program of Shenzhen (JCYJ20150831154213680) and Innovation project for graduate students of Nanchang University (CX201786). * indicates the co-corresponding author.

Y. Liu, Q. Liu, M. Zhang and Q. Yang are with the Department of Electronic Information Engineering, Nanchang University, Nanchang 330031, China. (liuyiling@email.ncu.edu.cn, {liuqiegen, zhangminghui}@ncu.edu.cn, QingxinYang@email.ncu.edu.cn). Y. Liu did the work during her internship at Paul C. Lauterbur Research Center for Biomedical Imaging, Chinese Academy of Sciences, Shenzhen, China.

S. Wang and D. Liang are with Paul C. Lauterbur Research Center for Biomedical Imaging and the Medical AI Research Center, Shenzhen Institutes of Advanced Technology, Chinese Academy of Sciences, Shenzhen 518055, China (sophiasswang@hotmail.com, dong.liang@siat.ac.cn).

researchers focus on learning specific projection in the traditional alternative model with deep convolutional neural network (CNN) [43] - [45]. Another direction is to form the model-based network with CNN-based regularization prior, in which the weights of CNN are shared across iterations/layers [46].

In this work, under the effective IFR-CS model, we form a new IFR-Net by inheriting merits of parameter-learning from ADMM-Net [41] and the CNN-weights learning from [46], [48] for CS-MRI reconstruction. Different from the weight-sharing CNN denoiser in [46], the weights of CNN in this network are all separately trained. With the fine structure preserving ability of IFR-CS, the proposed IFR-Net can also be trained with small image dataset while added some trainable CNN blocks. Concretely, this network is composed of three main modules: a reconstruction module for restoring an MR image from highly undersampled $k$-space, a CNN-based denoising module for removing noise-like artifacts caused by undersampling, and a feature refinement module for picking structure information from the residual image. Experimental results for a range of undersampling rates and different sampling patterns have shown that IFR-Net can achieve superior results with the preservation of more structural information, compared to the traditional iterative methods including the initial version IFR-CS and several recent deep learning approaches.

The main contributions are as follows:
- The present IFR-Net is a network developed from the IFR-CS scheme. Unlike most of model-driven networks derived from specific alternative schemes, IFR-Net is formed by unrolling the iterative feature refinement model to multiple layers/modules. Due to the excellent feature refinement property, IFR-Net can be trained with small image dataset.
- In IFR-Net, besides of the regularization parameter, the utmost feature refinement operator in IFR-CS also becomes trainable. Hence a fully parameter-learning strategy is presented. Furthermore, CNN-based inversion blocks are integrated into the denoising module of the network to increase the network capacity.

## II. PRELIMINARIES

### A. General CS-MRI

Assume $x \in \mathbb{C}^M$ is a MR image to be reconstructed and $y \in \mathbb{C}^N (N < M)$ is the undersampled $k$-space data. The general CS-MRI model can be formulated as follows:

$$\min_x \frac{1}{2}\|F_p x - y\|_2^2 + \sum_{l=1}^{L} \lambda_l \|D_l x\|_1, \quad (1)$$

where the first term is the data fidelity term in the $k$-space domain, and the second term is the regularization term. In addition, $F_p$ denotes the undersampled Fourier encoding matrix, and $D_l$ denotes a transform matrix for a filtering operation, e.g., a discrete wavelet transform or a discrete cosine transform (DCT). Parameter $\lambda_l$ determines the trade-off between these two terms. There are many ways to tackle this problem. Alternating direction method of multipliers (ADMM) is a widely utilized technique that has been proven to be efficient and generally applicable with a guaranteed convergence [47]. It considers the augmented Lagrangian function of a given CS-MRI model, and splits the variables into subgroups, which can be alternatively optimized by solving a few simple sub-problems. For instance, by introducing an auxiliary variable $u$, Eq. (1) can be rewritten as

$$\min_{x,u} \frac{1}{2}\|F_p x - y\|_2^2 + \sum_{l=1}^{L} \lambda_l \|D_l u\|_1 + \frac{\rho}{2}\|x - u\|_2^2, \quad (2)$$

where $\rho$ is the penalty parameter.

### B. IFR-CS Model

The basic idea of IFR-CS [21] is to design a linear feature-refining module to recover useful image details while removing the noise and noise-like artifacts. For Eq. (2), its IFR-CS formulation can be written as follows:

$$\begin{cases} x^{(n+1)} = \arg\min_x \frac{1}{2}\|F_p x - y\|_2^2 + \frac{\rho}{2}\|x - x_t^{(n)}\|_2^2 \\ u^{(n+1)} = \arg\min_u \frac{\rho}{2}\|x^{(n)} - u\|_2^2 + \sum_{l=1}^{L} \lambda_l \|D_l u\|_1 \\ x_t^{(n+1)} = u^{(n+1)} + T^{(n+1)} \otimes (x^{(n+1)} - u^{(n+1)}) \end{cases}, \quad (3)$$

where $T$ is the feature descriptor partly controlled by a parameter $V$ (Details can be seen in **Appendix A**), $x_t$ is the feature-refined denoised image, $n$ is the $n$-th iteration, $\otimes$ is a dot product operation, and $x^{(n+1)}$ can be efficiently computed by fast Fourier transform. As shown in Fig. 1, the IFR-CS approach consists of three main steps: a sparsity-promoting denoising step, a feature refinement step, and a Tikhonov regularization step. The feature descriptor $T$ in feature refinement step is similar to a special filter, which filters out unwanted signals of the residual image and keeps the wanted signals. Then, the wanted signals can be added back to the denoised image. This step can enable the denoised image to keep more details.

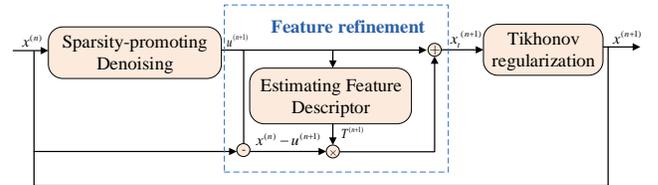

**Fig. 1.** Overview of the IFR-CS approach.

### C. CNNs-based Iterative Inversion Block

Over the past decades, most traditional methods have applied well-chosen filter-based regularized iterative algorithms to solve the ill-posed inversion problem in imaging processing. Although these iterative approaches can achieve a state-of-the-art performance, a high computational cost and fixed transform kernels put traditional iterative methods at a disadvantage as compared to deep learning based methods. To alleviate this deficiency, the researchers in [48] explored the relationship between conventional iterative inversion operations and the CNN framework. As shown in Fig. 2, the unfolded architecture of the iterative shrinkage procedure with a sparsifying transform ($\mathbf{x} = S_{\lambda/L}(\frac{1}{L}\mathbf{W}^*\mathbf{H}^*\mathbf{y} + (\mathbf{I} - \frac{1}{L}\mathbf{W}^*\mathbf{H}^*\mathbf{H}\mathbf{W})\mathbf{x}_0)$ [48]) is similar to the framework of a convolution network. We can thus utilize the powerful CNN to train a more suitable sparse operator for our CS-MRI reconstruction task without high computational costs during the reconstruction.

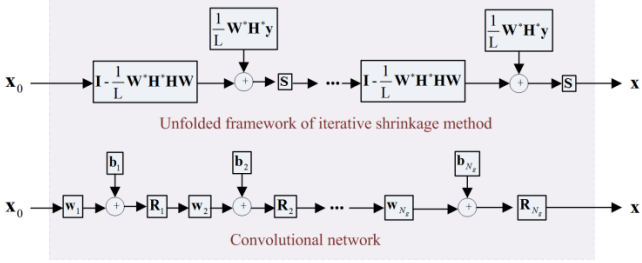

**Fig. 2.** Block diagrams for unfolded version of iterative shrinkage method with sparsifying transform and convolution network. Here, $L$ is the Lipchitz constant, $\mathbf{x}_0$ is the initial estimates, $\mathbf{W}$ and $\mathbf{H}$ are transforms, $\mathbf{W}^*$ and $\mathbf{H}^*$ are the corresponding transpose transforms of $\mathbf{W}$ with $\mathbf{H}$, $\mathbf{y}$ denotes the vector of the measurements, $\mathbf{S}$ is the soft-thresholding operator determined by the regularization parameter $\lambda$ and Lipchitz constant $L$, $\mathbf{b}_i$ is the learned bias, $\mathbf{w}_i$ is the learned convolutional kernel, $N_g$ is the number of network layers, and $\mathbf{R}_i$ denotes the Relu operator.

## III. PROPOSED IFR-NET MODEL

The proposed IFR-Net model contains the network design part and parameter training part. In particular, the network architecture is formed by two successive stages: an iterative scheme employing gradient descent to tackle the IFR-CS model, and unrolling the iterative scheme to a network with CNN layers. To facilitate the description of the present method, we tabulate the notations used hereafter in Table I.

### A. Proposed Network

First, we use a gradient descent to tackle the IFR-CS model and generate the iterative scheme. Differing from the IFR-CS, we directly employ the gradient descent algorithm to solve the first sub-problem in Eq. (3), which is thus modified as

$$\begin{cases} x^{(n+1)} = F^{-1}(\dfrac{y+\rho F(x_t^{(n)})}{FF_p^T F_p F^T + \rho I}) \\ u^{(n+1,k)} = (1-\rho l_r)u^{(n+1,k-1)} + \rho l_r x^{(n+1)} - \sum_{l=1}^{L} l_r \lambda_l D_l^T \left\| D_l u^{(n+1,k-1)} \right\|_1, \\ x_t^{(n+1)} = u^{(n+1)} + T^{(n+1)} \otimes (x^{(n+1)} - u^{(n+1)}) \end{cases} \quad (4)$$

where $l_r$ is the step size, $I$ denotes a unit matrix, $F_p^T F_p$ is a block circulant matrix, and $FF_p^T F_p F^T$ is a diagonal matrix consisting of ones and zeros with the ones corresponding to the sampled locations in $k$-space. In addition, $F$ denotes the normalized full Fourier encoding matrix satisfying $FF^T = 1_N$.

TABLE I SUMMARIZATION OF NOTATIONS.

| Notation | Description | Notation | Description |
|---|---|---|---|
| $y$ | Undersampled $k$-space | $c_1$ | First convolution layer in one CNN-based block |
| $u$ | Auxiliary variable that can be seen as a denoised image | $c_2$ | Second convolution layer in one CNN-based block |
| $x_t$ | Feature-refined denoised image | $h$ | Nonlinear layer |
| $x$ | Image to be reconstructed from undersampled $k$-space | $L$ | Filter number |
| $T$ | Feature descriptor | $w_f$ | Size of filters in $c_1$ |
| $\otimes$ | Dot product operation | $f$ | Size of filters in $c_2$ |
| $\lambda_l$ | Regularization parameter | $N_c$ | Number of positions |
| $I$ | Unit matrix | $w_1$ | $L$ filters with size of $w_f \times w_f$ in $c_1$ |
| $\rho$ | Penalty parameter | $b_1$ | $L$ - dimensional biases vector in $c_1$ |
| $D_l$ | Transform matrix for a sparsifying operation | $w_2$ | Filter with size of $f \times f \times L$ in $c_2$ |
| $l_r$ | Step size in gradient-descent algorithm | $b_2$ | One-dimensional bias vector in $c_2$ |
| $F_p^T F_p$ | Block circulant matrix | $\{p_i\}_{i=1}^{N_c}$ | Predefined positions uniformly located within $[-1,1]$ |
| $F$ | Normalized full Fourier encoding matrix | $\{q_i^{(n,k)}\}_{i=1}^{N_c}$ | Values at $\{h_i\}_{i=1}^{N_c}$ positions for $k$-th block in $n$-th stage. |
| $N_s$ | Stage number | $S_{PLF}(\cdot)$ | Piecewise linear function determined by a set of control points $\{p_i, q_i^{(n,k)}\}_{i=1}^{N_c}$ |
| $K$ | CNN-based block number | $\Theta$ | Parameters in the network |
| $\mu_1$ | $1-\rho l_r$ | $\hat{x}(y,\Theta)$ | Output of the network |
| $\mu_2$ | $\rho l_r$ | $x^{gt}$ | Ground-truth image |
| $*$ | Convolution operation | $\psi = \{y, x^{gt}\}$ | Pair of training data |

Second, we unroll the iterative scheme Eq. (4) by incorporating a CNN layer to form a network. In particular, to improve the reconstruction performance and increase the network capacity, we utilize CNN-based iterative inversion blocks to achieve the $\sum_{l=1}^{L} l_r \lambda_l D_l^T \|D_l u^{(n+1,k-1)}\|_1$ operation in Eq. (4). Based on the iterative process in Eq. (4), we propose a network-based CS-MRI method composed of a fixed number of phases, each of which strictly corresponds to an iteration in traditional IFR-CS.

$$\begin{cases} \mathbf{X^{(n+1)}}: x^{(n+1)} = F^{-1}\left(\dfrac{y + \rho^{(n+1)} F(x_t^{(n)})}{FF_p^T F_p F^T + \rho^{(n+1)} I}\right) \\ \mathbf{Z^{(n+1)}}: \begin{cases} u^{(n+1,k)} = \mu_1^{(n+1,k)} u^{(n+1,k-1)} + \mu_2^{(n+1,k)} x^{(n+1)} - c_2^{(n+1,k)} \\ c_1^{(n+1,k)} = w_1^{(n+1,k)} * u^{(n+1,k)} + b_1^{(n+1,k)} \\ h^{(n+1,k)} = S_{PLF}(c_1^{(n+1,k)}; \{p_i, q_i^{(n+1,k)}\}_{i=1}^{N_c}) \\ c_2^{(n+1,k)} = w_2^{(n+1,k)} * h^{(n+1,k)} + b_2^{(n+1,k)} \end{cases} \\ \mathbf{R^{(n+1)}}: x_t^{(n+1)} = u^{(n+1)} + T^{(n+1)} \otimes (x^{(n+1)} - u^{(n+1)}) \end{cases}$$
(5)

where $n \in [1,2,\cdots,N_s]$ is the number index of stages, $k \in [1,2,\cdots,K]$ is number index of the block, $\mu_1 = 1 - \rho l_r$, $\mu_2 = \rho l_r$, $*$ denotes the convolution operation. $w_1^{(n+1,k)}$ denotes $L$ filters with a size of $W_f \times W_f$, which roughly takes the place of $D$, and $b_1^{(n+1,k)}$ is an $L$-dimensional biases vector. $w_2^{(n+1,k)}$ corresponds to a filter with a size of $f \times f \times L$, which can be seen as the $D^T$ in $\sum_{l=1}^{L} l_r \lambda_l D_l^T \|D_l u^{(n+1,k-1)}\|_1$, and $b_2^{(n+1,k)}$ is a one-dimensional bias vector. $S_{PLF}(\cdot)$ is a piecewise linear function similar to the function of Relu layer in the CNN, which is used to achieve the shrinkage function of $\|\cdot\|_1$ in $\sum_{l=1}^{L} l_r \lambda_l D_l^T \|D_l u^{(n+1,k-1)}\|_1$. It is determined by a set of control points $\{p_i, q_i^{(n+1,k)}\}_{i=1}^{N_c}$, $\{p_i\}_{i=1}^{N_c}$ are predefined positions uniformly located within $[-1,1]$, and $\{q_i^{(n+1,k)}\}_{i=1}^{N_c}$ are the trainable values at these positions for the $k$-th block at the $(n+1)$-th stage. Specially, assume variable $\alpha$ is the input of $S_{PLF}(\cdot)$, its output is:

$$S_{PLF}(\alpha) = \begin{cases} \alpha + q_1 - p_1, & \alpha < p_1, \\ q_r + (\alpha - p_r)(q_{r+1} - q_r) / (p_{r+1} - p_r), & p_1 \le \alpha \le p_{N_c}, \\ \alpha + q_{N_c} - p_{N_c}, & \alpha > p_{N_c}, \end{cases}$$

where $r = \left\lfloor \dfrac{\alpha - p_1}{p_2 - p_1} \right\rfloor$.

As can be seen in Eq. (5), at the $n$-th stage of the graph in the present IFR-Net, there are three types of nodes mapped from three types of operations, i.e., a reconstruction operation ($\mathbf{X}^{(n)}$), a sparsity-promoting denoising operation ($\mathbf{Z}^{(n)}$), and a feature refinement operation ($\mathbf{R}^{(n)}$). A whole data flow graph is a multiple repetition of the above stages corresponding to successive iterations, as depicted in Fig. 3. As seen in Fig. 3 (a), the proposed IFR-Net consists of several stages, where each stage is a component of three modules: a reconstruction module (**X**), a sparsity-promoting denoising module (**Z**) (the details of which are shown in Fig. 3 (b)), and a feature refinement module (**R**). It is also clear to see that our network can be regarded as the concatenation of many stages, and in each stage, and there are three types of modules inside. To concisely describe the network, we take the $n$-th stage as an example to depict the modules of the network as follows:

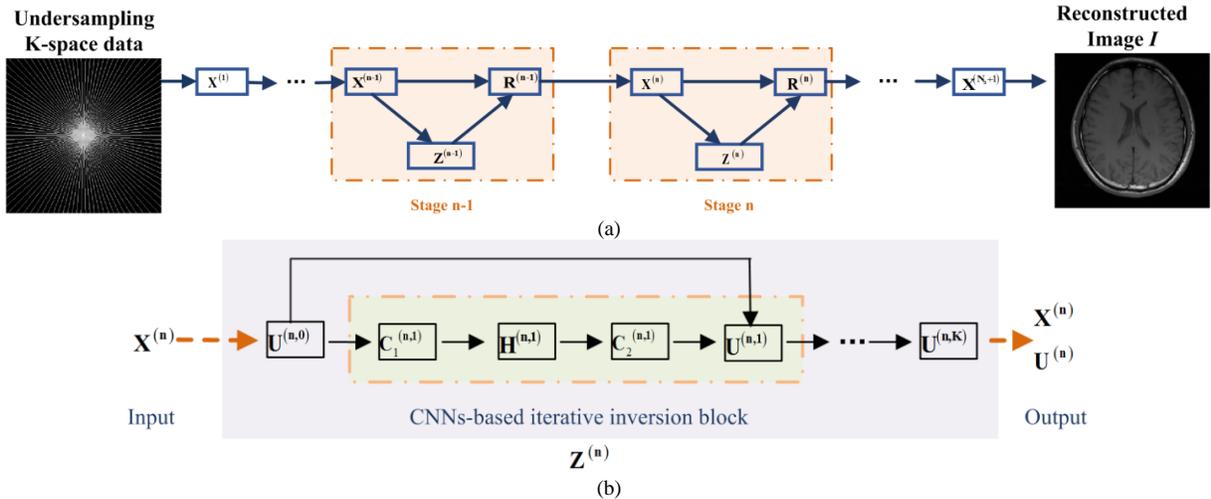

**Fig. 3.** Framework of the proposed IFR-Net.

*1) Reconstruction module ( $X^{(n)}$ )*

The inputs of this module are $y$ and $x_t^{(n-1)}$. The output of this module is defined as

$$x^{(n)} = F^{-1}(\frac{y + \rho^{(n+1)} F(x_t^{(n)})}{FF_p^T F_p F^T + \rho^{(n+1)} I}), \tag{6}$$

It should be noted that, in the first stage ($n=1$), $x_t^{(0)}$ is initialized as zero, and therefore $x^{(1)} = F^{-1}(\frac{y}{FF_p^T F_p F^T + \rho^{(n)} I})$.

*2) Sparsity-promoting denoising module ( $Z^{(n)}$ )*

To learn a more suitable sparse operator for a CS-MRI reconstruction task and increase the network capacity, we introduce several CNN-based convolution layers into this module. As shown in Fig. 3 (b), the module is composed of several CNN-based blocks, and in each block there are four layers: a united layer $U^{(n,k)}$, convolution layers $C_1^{(n,k)}$ and $C_2^{(n,k)}$, and a nonlinear transform layer $H^{(n,k)}$.

*United layer ( $U^{(n,k)}$ ):*

The inputs of this layer are $u^{(n,k-1)}$, $x^{(n)}$, and $c_2^{(n,k)}$. The resulting output is

$$u^{(n,k)} = \mu_1^{(n,k)} u^{(n,k-1)} + \mu_2^{(n,k)} x^{(n)} - c_2^{(n,k)}, \tag{7}$$

Note that, in the first block, $u^{(n,1)} = x^{(n)}$.

*Convolution layer ( $C_1^{(n,k)}$ ):*

The input of this layer is $x^{(n)}$. The output of this layer is

$$c_1^{(n,k)} = w_1^{(n,k)} * u^{(n,k)} + b_1^{(n,k)}, \tag{8}$$

*Nonlinear transform layer ( $H^{(n,k)}$ ):*

The input of this layer is $c^{(n,k)}$. The output of this layer is

$$h^{(n,k)} = S_{PLF}(c_1^{(n,k)}; \{p_i, q_i^{(n,k)}\}_{i=1}^{N_c}), \tag{9}$$

*Convolution layer ( $C_2^{(n,k)}$ ):*

Given the input $h^{(n,k)}$, the output of this layer is defined as

$$c_2^{(n,k)} = w_2^{(n,k)} * h^{(n,k)} + b_2^{(n,k)}, \tag{10}$$

*3) Feature refinement module ( $R^{(n)}$ )*

The inputs of this module are $u^{(n)}$, $x^{(n)}$. The output is

$$x_t^{(n)} = u^{(n)} + T^{(n)} \otimes (x^{(n)} - u^{(n)}), \tag{11}$$

*B. Network Training*

In the network training procedure, we aim to learn the following parameters $\Theta$ in this deep architecture: $\rho^{(n)}$ in the reconstruction module; filters $w_1^{(n,k)}$, $w_2^{(n,k)}$ and biases $b_1^{(n,k)}$, $b_2^{(n,k)}$ in two convolution layers in the sparsity-promoting denoising module, along with $\mu_1^{(n,k)}$, $\mu_2^{(n,k)}$, and $\{q_i^{(n)}\}_{i=1}^{N_c}$; and $V^{(n)}$ in feature extraction operator $T^{(n)}$ in the feature refinement module. Based on the parameters $\Theta$, we set the undersampled *k*-space $y$ as the input and $\hat{x}(y, \Theta)$ is the network output. By updating the parameters $\Theta$, the training procedure is devoted to making the output $\hat{x}(y, \Theta)$ close to the ground-truth image $x^{gt}$ as soon as possible. For the given pairs of training data $\psi = \{y, x^{gt}\}$ and $\Theta$, we set the normalized mean square error (NMSE) as the loss function, i.e.,

$$E(\Theta) = \frac{1}{|\psi|} \sum_{(y, x^{gt}) \in \psi} \sqrt{\|\hat{x}(y, \Theta) - x^{gt}\|_2^2} / \sqrt{\|x^{gt}\|_2^2}, \tag{12}$$

where $|\psi|$ denotes the number of data pairs in $\psi$. The stochastic gradient descent (SGD) algorithm is used to optimize these parameters by minimizing the NMSE loss. The gradients of the loss function $E(\Theta)$ of the restored result with parameters $\Theta$ are computed with the back-propagation (BP) algorithm. This procedure is detailed in **Appendix B**.

IV. EXPERIMENT RESULTS

To demonstrate the effectiveness of IFR-Net, we compare it with the initial version of our previous model IFR-CS [21] and a representative model-based network, ADMM-Net [41], for real-valued experiments. For complex-valued testing, we firstly make a simple comparison of IFR-Net and IFR-CS with 3 brain images from our scanner. Then, to make it more convincing, further comparisons among IFR-Net and patch-based algorithm PANO [16], dictionary learning method FDLCP [13] and data-driven network D5-C5 [30] are conducted on 50 knee images from the open dataset *FastMRI*[2] [50]. For a fair comparison, the parameters of IFR-CS, PANO and FDLCP are hand-tuned for the best performance. As for ADMM-Net method, we initialize the parameters as the default with 15 stages, and the filters in each stage are set to be eight 3×3 DCT bases. Both ADMM-Net and our proposed method are trained using the same 100 brain images. All of these experiments are implemented in MATLAB 2017a with MatConvNet toolbox on a PC equipped with an Intel(R) Xeon (R) CPU X5690 @ 3.47 GHz. The peak signal-to-noise ratio (PSNR), high-frequency error norm (HFEN), and structural similarity (SSIM) are used to quantitatively evaluate the qualities of the reconstruction results. For convenient reproducibility, a demonstration code can be downloaded from the following website: https://github.com/yqx7150/IFR-Net-Code.

*A. Data Acquisition*

**Dataset 1.** For real-valued MR reconstruction, we train and test our network using 100 and 5 brain amplitude images, respectively. For the complex-valued experiments, we train our IFR-Net with 100 brain images. All of these data are scanned from a 3T Siemens MAGNETOM Trio scanner using the mixed-weighted (T1, T2, and Proton Density-PD) turbo spin echo sequence. For T1- and T2-weighted data, the field of view (FOV) is 220 mm × 220 mm, and the slice thickness is 0.86 mm. For PD-weighted data, the field of view (FOV) is 220 mm × 220 mm and the slice thickness is 1.06 mm. In the real-valued reconstruction comparison, we obtain the real valued and single channel images using the sum-of-square (SOS) operator. For complex-valued data, we utilize the adaptive coil-combine method [49] to obtain complex single-channel images. The top line of Fig. 4 shows the real valued testing images used in our experiments.

**Dataset 2.** In this dataset, we select 400 knee data from the "*single coil train*" set of *FastMRI* as training set and select 50

---
[2] https://fastmri.med.nyu.edu

knee data from "*single coil val*" set of *FastMRI* as testing set. All of these data are cropped to the center $320 \times 320$ region. As the data we downloaded are complex single-channel *k*-space data, we use them for the complex experiments. For real-valued experiments, we also utilize the SOS operator to obtain the real-valued single channel images. Some testing images are shown in the bottom line of Fig. 4.

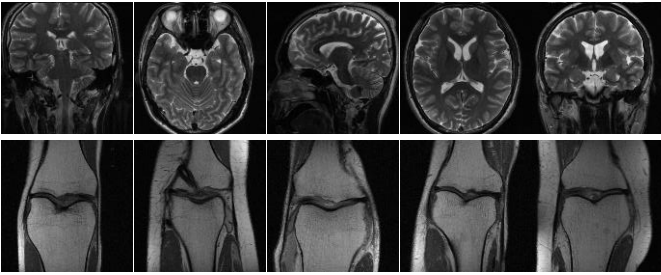

**Fig. 4.** Testing images. Top: Five real valued testing brain images in dataset 1. Bottom: Five examples of amplitude knee images in dataset 2.

**Sampling Masks.** Three different types of undersampling patterns are tested, i.e., 1D random, 2D random, and pseudo radial sampling. For the pseudo radial mask, 10%, 20%, 30%, and 40% retained raw k-space data are simulated, representing 10×, 5×, 3.3×, and 2.5× accelerations, respectively. For 1D random, 2D random, and pseudo radial masks, 25% retained *k*-space data are simulated. A visualization of some sampling masks is depicted in Fig. 5.

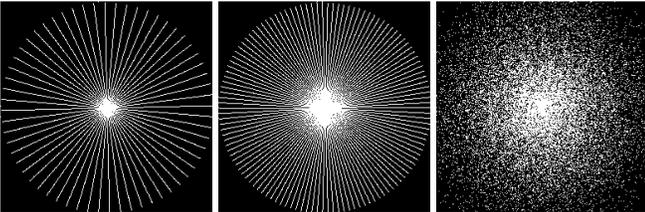

**Fig. 5.** Sampling masks used in the experiments. Left: Pseudo radial sampling of 10%. Middle: Pseudo radial sampling of 20%. Right: 2D random sampling of 25%.

### B. Network Setup

In this subsection, we investigate the effects of different depths and widths on the proposed IFR-Net. The comparisons are conducted by varying one parameter at a time while keeping the rest fixed at their nominal values and they are all trained and tested on dataset 1.

**Depth.** There are two factors influencing the depth of the net: the numbers of stages and iterative inversion blocks. Initially, we train deeper networks by increasing the number of stages using 50 real-valued brain images. Fig. 6 illustrates the average PSNR, HFEN, and SSIM values when varying the number of stages under the pseudo radial sampling rate of 40%. Other network parameters are set as $w_f = 3$, $L = 8$, and $K = 2$, and the filters are initiated with eight 3×3 DCT basis. It can be seen that the average PSNR, HFEN, and SSIM improve with the increasing number of stages for $N < 9$. However, in the case of $N \geq 9$, the values fluctuate sharply. Hence, it can be concluded that deeper networks achieve better results; however deeper is not always better because the gradients may vanish with very deep networks.

We then train the IFR-Net using numbers of iterative inversion blocks of $K = 2, 3, 4$. The average PSNR, HFEN, and SSIM values are provided in Table II. We can see from the table that the PSNR is improved by approximately 0.07 dB, and the SSIM is improved by approximately 0.006 when increasing the number of blocks by 1. However, the HFEN values are nearly the same when $K = 3$ and 4, and the performance is even a little better when $K = 3$. We can conclude that the increasing number of iterative blocks has a small contribution to improving the reconstruction performance.

**Width.** To test the effects of the network width, we prepare the networks for training using filter numbers of 8, 64, and 128, respectively. Besides of setting $N = 7$, $K = 2$, and $w_f = 3$, here we randomly initialize the filters, due to that the DCT basis has a limitation regarding the numbers of filters. More details on the filter initiation are provided in the discussion section.

As shown in Table III, the PSNR, HFEN, and SSIM achieve the highest values simultaneously when the IFR-Net network is trained using 64 filters.

TABLE II AVERAGE PSNR, HFEN, AND SSIM RESULTS OF DIFFERENT NUMBERS OF BLOCKS.

| Block Number | $K = 2$ | $K = 3$ | $K = 4$ |
|---|---|---|---|
| PSNR/dB | 36.1146 | 36.1841 | **36.2422** |
| HFEN | 0.4410 | **0.4275** | 0.4284 |
| SSIM | 0.9405 | 0.9411 | **0.9418** |

TABLE III AVERAGE PSNR, HFEN, AND SSIM RESULTS OF DIFFERENT NUMBERS OF FILTERS.

| Filter Number | $L = 8$ | $L = 64$ | $L = 128$ |
|---|---|---|---|
| PSNR/dB | 35.9539 | **36.1740** | 35.9918 |
| HFEN | 0.4449 | **0.4237** | 0.4357 |
| SSIM | 0.9394 | **0.9411** | 0.9395 |

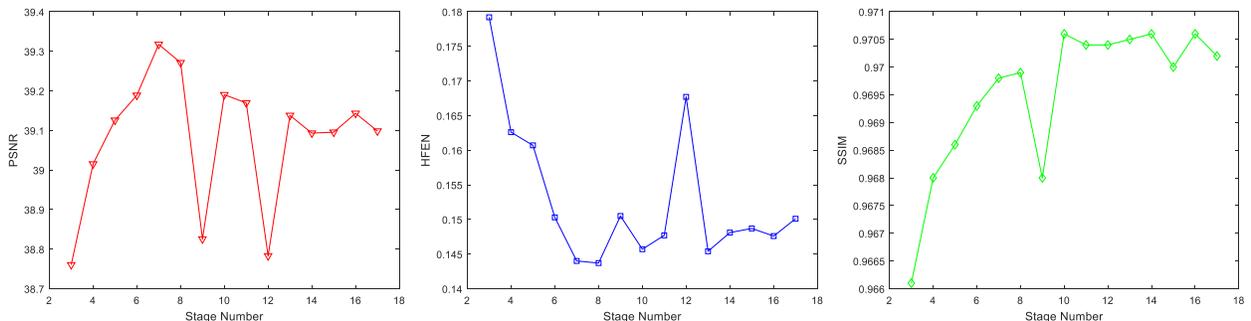

**Fig. 6.** Average PSNR, HFEN, and SSIM values versus stage number. The experiment was conducted using pseudo radial sampling of 40%.

## C. Performance Comparison on Real-valued Data

**Reconstruction under Different Sampling Rate.** Table IV tabulates the quantitative comparison results of the traditional IFR-CS and state-of-the-art ADMM-net with pseudo radial sampling rates of 10%, 20%, 30%, and 40%. It can be obviously observed that IFR-Net achieves the best scores for all of the quantitative comparisons. In particular, although ADMM-Net obtains better values than IFR-CS with 10% sampled data, IFR-CS obtains better average results when the sampling rate is $\geq 20\%$. The top and middle lines of Fig. 7 show visual reconstruction examples on the testing brain images using IFR-CS, ADMM-Net, and IFR-Net from the simulated $k$-space data with 10% and 20% pseudo radial sampling masks, respectively. We can clearly see from the enlargements and reconstruction errors in Fig. 7 that IFR-Net effectively suppresses most of the artifacts and preserves the image details, whereas the competing methods show an obvious loss of structural details. For the 20% sampled data in Fig. 8, both IFR-Net and the competing methods obtain fine reconstruction details without a significant loss in structural information, whereas it can still be observed from the enlargements and errors that IFR-Net achieves a more accurate reconstructed line profile and structural details.

**Reconstruction under Different Sampling Mask.** This subsection presents the comparison results of different sampling masks (1D random, 2D random, and pseudo radial masks with a sampling rate of 25%). The quantitative results are tabulated in Table V. Clearly, the results indicate that the proposed method obtained the best performances with various sampling patterns. Similarly, for a visual comparison, we present example reconstruction results in the bottom line of Fig. 7. Despite slight differences between the reconstructed images and the reference even in the enlargement part, the reconstruction errors show that IFR-Net outperforms the comparison methods with less structural loss, particularly in the edges of the organ.

## D. Performance Comparison on Complex-valued Data

In general, MR images are complex-valued, and their phase information is also important. In this part, we provide several reconstruction examples of complex-valued MR images. The proposed IFR-Net can achieve the desired result even with complex-valued data.

**Comparison on Complex-valued Dataset 1.** At first, we compared the proposed IFR-Net to its initial version IFR-CS with 25% 2D-random sampling and 20% pseudo radial sampling masks on dataset 1. For quantitative values, it can be observed from Table VI that IFR-Net achieves better results than IFR-CS for all evaluation indexes. In particular, the proposed method outperforms IFR-CS by 3.0991 dB under a 20% pseudo radial sampling.

In terms of visual comparison, amplitude images of the complex-valued images reconstructed from 25% 2D random sampling and 20% pseudo radial $k$-space data are shown in Fig. 8. The enlargement and reconstruction errors show that our method outperforms IFR-CS in both artifact removal and the preservation of fine structures.

**Comparison on Complex-valued Open Dataset 2.** To further demonstrate the effectiveness of the proposed method, we conduct an extra comparison on dataset 2, in which the IFR-Net is compared to more state-of-the-art methods: the traditional algorithms PANO [16] and FDLCP [13], and the data driven network D5-C5 [30]. In this experiment, D5-C5 is trained with 400 single-channel complex images from dataset 2 and all of the methods are tested with 50 single-channel complex knee images selected from *FastMRI*. And they are also sampled with 25% 2D-random and 20% pseudo radial masks, respectively.

Table VII lists the average quantitative values of 50 images. IFR-Net outperforms the competing algorithms in terms of lower HFEN, higher PSNR and SSIM values. Fig. 9 visually compares two different slices at 25% 2D random and 20% pseudo radial sampling. It is evident from the error images that the reconstruction quality by deep learning approaches D5-C5 and IFR-Net are better than the traditional iterative methods PANO and FDLCP. Moreover, IFR-Net produces more visually pleasuring result than D5-C5 in the case of 20% pseudo radial sampling.

In addition, a comparison between our method and U-Net baseline from [50] on the undersampling schemes from the leaderboard of *FastMRI* is also added to better verify the effectiveness of IFR-Net. In this experiment, the U-Net is trained and validated by 300 files (about 10800 images) and 20 files (about 760 images) randomly selected from "*single coil train*" set of *FastMRI*, respectively. The results are tabulated in Table VIII. It is obvious that IFR-Net achieves better quantitative results at both 4-fold and 8-fold undersamplings.

TABLE VII AVERAGE PSNR, HFEN AND SSIM RESULTS OF PANO, FDLCP, VN, D5-C5 AND IFR-NET ON 50 COMPLEX-VALUED KNEE IMAGES.

| Methods | PANO | FDLCP | D5-C5 | IFR-Net |
|---|---|---|---|---|
| **25% 2D Random** | 32.5288 | 28.5380 | 33.4894 | **34.3510** |
| | 0.8835 | 1.6910 | 0.7533 | **0.6119** |
| | 0.7604 | 0.6020 | 0.7964 | **0.8378** |
| **20% Pseudo Radial** | 32.0404 | 28.8161 | 33.0088 | **33.9234** |
| | 1.0217 | 1.7993 | 0.8609 | **0.7021** |
| | 0.7344 | 0.6041 | 0.7743 | **0.8195** |

TABLE VIII
AVERAGE PSNR, HFEN, and SSIM RESULTS OF U-NET BASELINE AND IFR-NET.

| Methods | | PSNR/dB | HFEN | SSIM |
|---|---|---|---|---|
| **4-fold** | U-Net | 29.4116 | 1.5600 | 0.7785 |
| | **IFR-Net** | **30.8883** | **1.1024** | **0.8043** |
| **8-fold** | U-Net | 25.8614 | 2.1253 | 0.6846 |
| | **IFR-Net** | **27.1429** | **1.7510** | **0.6924** |

Table IV AVERAGE PSNR, HFEN, AND SSIM RESULTS OF IFR-CS, ADMM-NET, AND IFR-NET WITH SAMPLING RATES OF 10%, 20%, 30%, AND 40%.

| Rates | PSNR/dB | | | HFEN | | | SSIM | | |
|---|---|---|---|---|---|---|---|---|---|
| | IFR-CS | ADMM-Net | IFR-Net | IFR-CS | ADMM-Net | IFR-Net | IFR-CS | ADMM-Net | IFR-Net |
| **10%** | 28.9242 | 29.1359 | **30.0663** | 1.7269 | 1.7218 | **1.5103** | 0.8109 | 0.8127 | **0.8390** |
| **20%** | 33.6998 | 32.7612 | **34.8249** | 0.6628 | 0.8856 | **0.6497** | 0.9130 | 0.8939 | **0.9280** |
| **30%** | 36.2397 | 36.1112 | **37.8444** | 0.3365 | 0.4013 | **0.2926** | 0.9449 | 0.9439 | **0.9571** |
| **40%** | 38.1590 | 37.6589 | **39.5642** | 0.1505 | 0.2135 | **0.1377** | 0.9627 | 0.9585 | **0.9703** |

TABLE V  Average PSNR, HFEN and SSIM results of IFR-CS, ADMM-Net, and IFR-Net with different sampling patterns.

| Methods | 1D-Random | | | 2D-Random | | | Pseudo Radial | | |
|---|---|---|---|---|---|---|---|---|---|
| | PSNR/dB | HFEN | SSIM | PSNR/dB | HFEN | SSIM | PSNR/dB | HFEN | SSIM |
| IFR-CS | 27.7484 | 1.4519 | 0.8286 | 34.8108 | 0.4092 | 0.9291 | 35.2254 | 0.4566 | 0.9336 |
| ADMM-Net | 28.6608 | 1.4172 | 0.8375 | 34.0650 | 0.6214 | 0.9139 | 35.6067 | 0.4742 | 0.9374 |
| IFR-Net | **29.0481** | **1.3868** | **0.8472** | **36.2627** | **0.4051** | **0.9421** | **36.7340** | **0.4277** | **0.9465** |

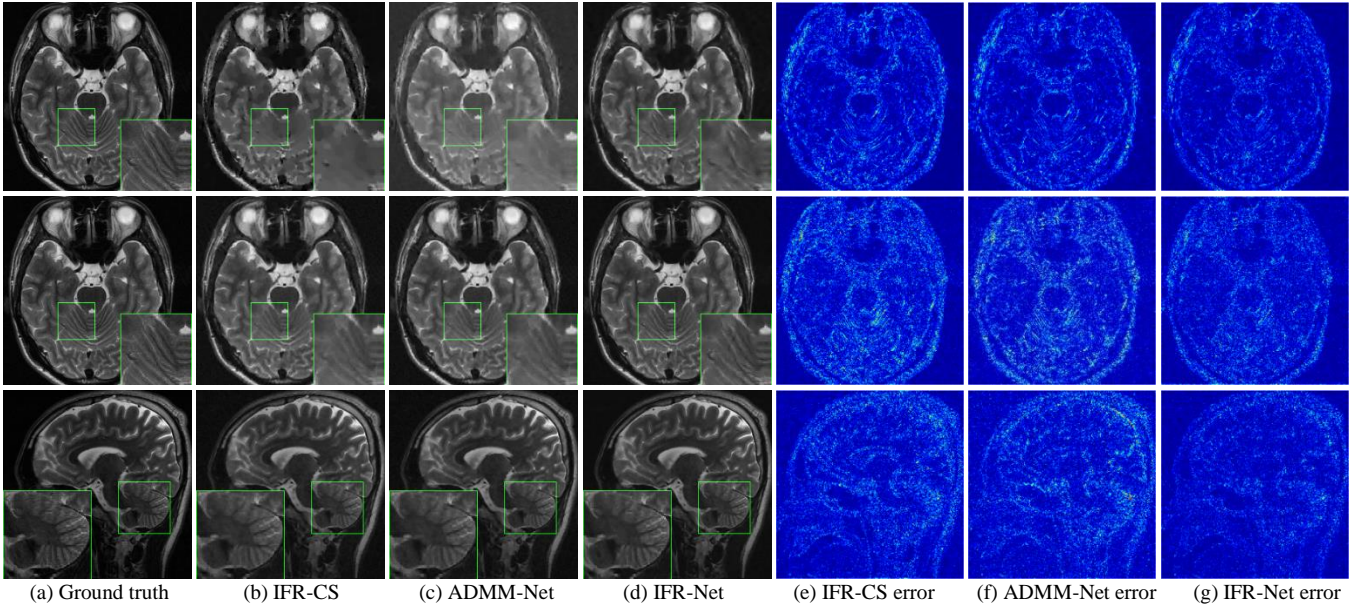

(a) Ground truth  (b) IFR-CS  (c) ADMM-Net  (d) IFR-Net  (e) IFR-CS error  (f) ADMM-Net error  (g) IFR-Net error

**Fig. 7.** Real-valued reconstruction results on brain image. Top: Reconstruction from 10% pseudo radial sampling. Middle: Reconstruction from 20% pseudo radial sampling. Bottom: Reconstruction from 25% 2D random sampling.

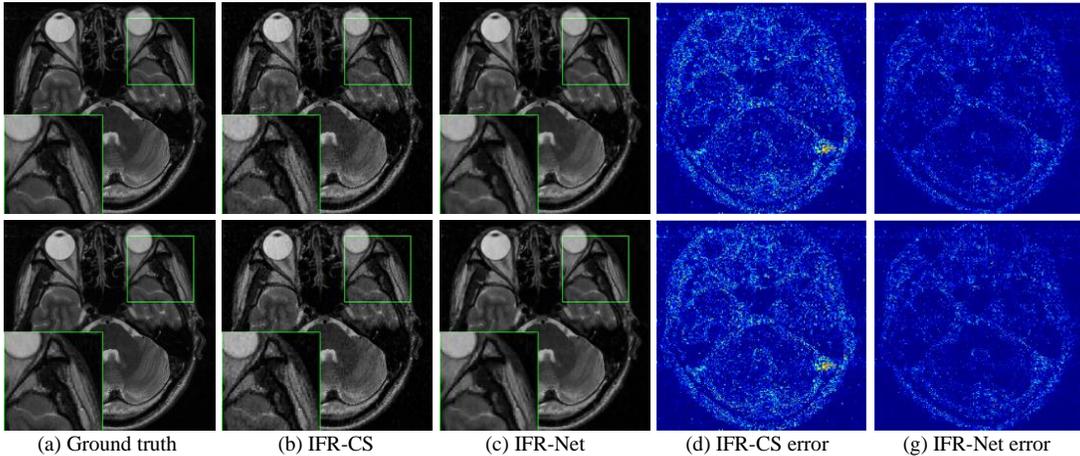

(a) Ground truth  (b) IFR-CS  (c) IFR-Net  (d) IFR-CS error  (g) IFR-Net error

**Fig. 8.** Complex-valued reconstruction results on brain images. Top: Reconstruction from 25% 2D random sampling. Bottom: Reconstruction from 20% pseudo radial sampling.

## IV. DISCUSSION

**Filter Initialization.** There are two types of initialization of the filters in sparsity-promoting denoising module: model-based initialization and random initialization. In a model-based case, the initialized filter is an effective and widely used sparse transform operator, e.g., DCT bases. The filters used can then be further learned through network training, and thus become a more suitable transform operator for a specific task. In the case of a random initialization, the filters are generally initialized based on random values in a Gaussian distribution, similar to classical deep learning methods. Table IX provides example results of a network trained using the same setup with the exception of the initialized filter types. It is clear that the network initiated using the model-based filters achieves a better performance for all quantitative values. Nevertheless, there is a drawback to the DCT-based initialization in that the numbers of filters in the DCT bases are associated with the filter size, e.g., eight filters with a size of $3\times3$ (the first DCT basis is discarded), 24 filters with a size of $5\times5$, or 48 filters with a size of $7\times7$, etc [41][30]. Thus, if a task needs to use a large number of filters of a small size, we have to choose a random initialization of the filters. In this work, to achieve higher reconstruction accuracy while maintaining a fast computational speed, we take eight DCT bases with a size of $3\times3$ as the filter initialization in our experiments.

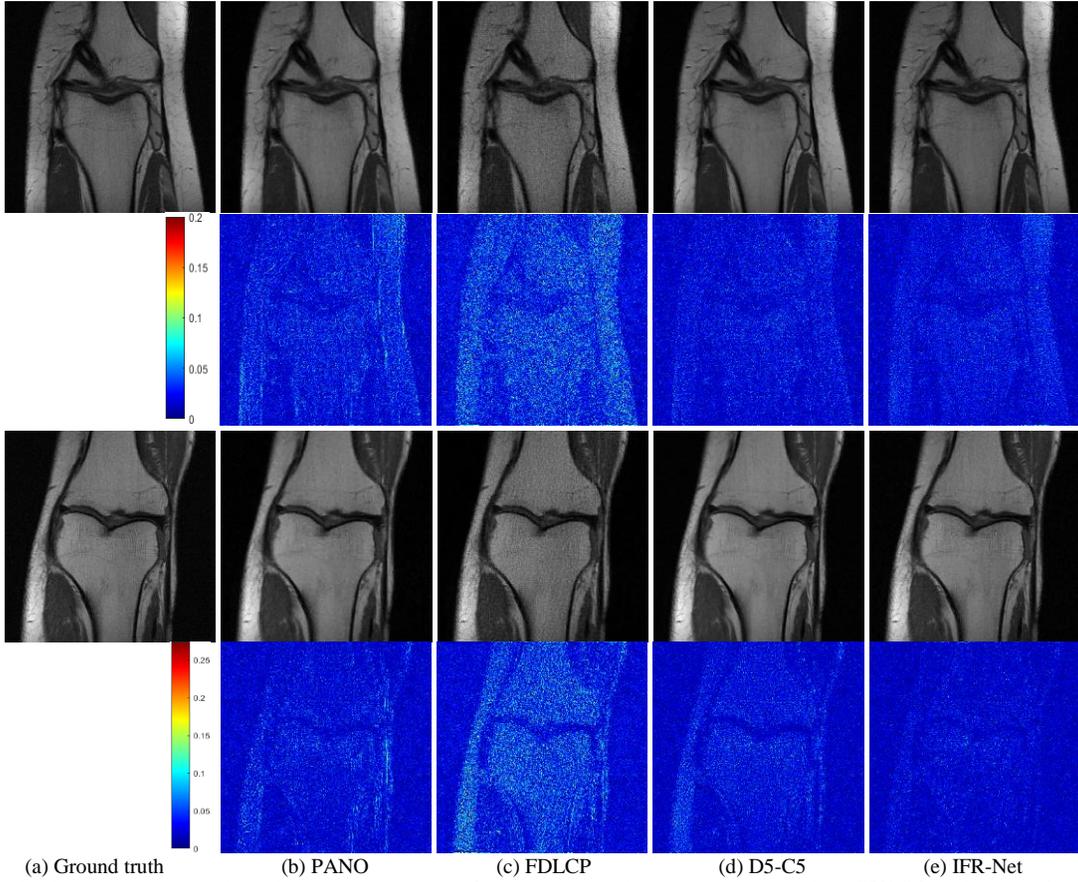

(a) Ground truth     (b) PANO     (c) FDLCP     (d) D5-C5     (e) IFR-Net

**Fig. 9.** Complex-valued reconstruction results on knee image. Top two: Reconstruction and corresponding error from 25% 2D random sampling. Bottom two: Reconstruction and corresponding error from 20% pseudo radial sampling.

TABLE VI
QUANTITATIVE RESULTS OF IFR-CS AND IFR-NET WITH 25% 2D-RANDOM AND 20% PSEUDO RADIAL SAMPLING MASKS ON 3 BRAIN IMAGES.

| Methods | | PSNR/dB | HFEN | SSIM |
|---|---|---|---|---|
| 25% 2D Random | IFR-CS | 30.1832 | 1.1406 | 0.8264 |
| | IFR-Net | **32.9585** | **0.8070** | **0.8834** |
| 20% Pseudo Radial | IFR-CS | 29.3025 | 1.4707 | 0.8001 |
| | IFR-Net | **32.1120** | **0.9839** | **0.8669** |

TABLE IX
AVERAGE PSNR, HFEN, and SSIM RESULTS OF DIFFERENT INITIALIZATIONS.

| Initialization | PSNR/dB | HFEN | SSIM |
|---|---|---|---|
| DCT | **36.1146** | **0.4410** | **0.9405** |
| Random | 35.9539 | 0.4449 | 0.9394 |

**Convergence Speed.** During the training phase, the training/validation error curve is used to detect the convergence speed of the network. This means that the network training gradually converges when the loss between two iterations is increasingly smaller, and the convergence of the error also indicates that the network has been trained well. Fig. 10 depicts a training/validation error curve for a network trained by 100 brain images with 25% 2D-random sampling. It can be observed that, as the number of iteration increases, the curve of both the training and validation error gradually converges to a low point. Although there are slight fluctuations in the iteration process, the overall network development trend gradually converges. In particular, we can easily see from the error curve that the proposed network converged at approximately the 130-th iteration. Because the training procedure of our network is driven by not only the data but also the traditional iterative algorithm model, the network is trained with a fast convergence speed.

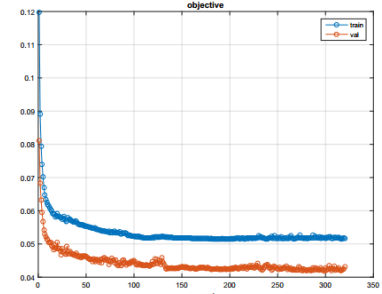

**Fig. 10.** Error curve in the training process.

**Generalization.** To evaluate the generalization capability of the proposed network, we apply an IFR-Net network learned from 100 brain images for the testing of knee images. Fig. 11 shows a reconstruction example of ADMM-Net and our IFR-Net with a pseudo radial sampling rate of 10%. We can see from the enlargement of the restored image that IFR-Net network achieves better reconstruction accuracy and preserves relatively fine structural details. This result indicates that the proposed network achieves a comparable or even better performance compared to other model-based deep network with less structure losses.

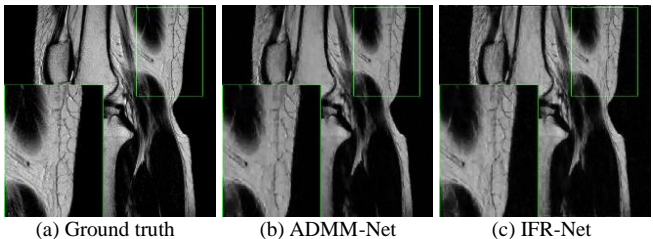

(a) Ground truth    (b) ADMM-Net    (c) IFR-Net

**Fig. 11.** Reconstruction results for a knee image with a pseudo radial sampling rate of 10%.

**Feature Descriptor.** In the initial version of IFR-CS, the feature refinement module plays an important role in picking up the structural information from the residual image. Particularly, the feature descriptor $T$ is the key map that affects the role of feature refinement module. Fig. 12 provides a visual example of feature descriptor maps under varying stages in our network. It can be seen that the details in a map of a latter stage are less than those in the former stage. The reason may be that, with an increase in the number of stages, the reconstructed image is better optimized and the amount of information loss in the denoising module decreases. Thus, there is not much structural information needed to be picked up. This may also indicate that our CNN-based inversion blocks in the denoising module are effective in denoising while preserving the structures well.

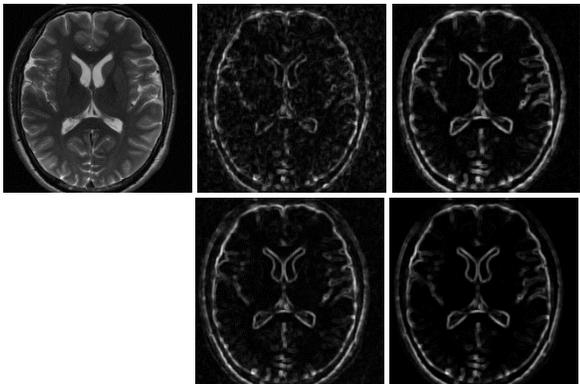

**Fig. 12.** Reference and corresponding feature descriptor maps in the feature refinement module for 2, 3, 7, and 9 stages under 25% 2D-random sampling.

**Extension for Multi-channel Images.** When dealing with multi-channel images, the input of the network can be formed by stacking the components along the channel direction, hence each channel can share the same parameters in the forward procedure. Except the variables at network input and output, the network structure for multi-channel images is the same as the single-channel counterpart. Alternate strategies In addition, the network can be also extended to multi-channel with the strategy similar to [46]. Specifically, the reconstruction module for multi-channel reconstruction can be modified with conjugate gradient (CG) included in the loop as in [46]. We will conduct the extension to multi-channel reconstruction in future study.

**Weights Sharing in Sparsity-promoting Module.** To decrease the demand for training data and training time, we conducted an experiment on training the IFR-Net with weights sharing in sparsity-promoting denoising module. In this experiment, the stage number is set as 5 and inner block as 2. The sampling trajectory is 40% pseudo radial sampling. As shown in Table X, the formal IFR-Net has a much better performance than the IFR-Net with weights sharing. Therefore, although the weights sharing can reduce the training data and training time, the weights trained separately for our network can boost the performance of it.

TABLE X
AVERAGE PSNR, HFEN, and SSIM RESULTS OF DIFFERENT WEIGHTS UPDATING WAY.

| Methods | PSNR/dB | HFEN | SSIM |
|---|---|---|---|
| IFR-Net-sharing | 33.2147 | 0.6704 | 0.9023 |
| **IFR-Net** | **39.2586** | **0.1458** | **0.9692** |

## V. CONCLUSION

In this paper, we presented a sparse operator-enhanced iterative network for fast CS-MRI reconstruction. Compared to the initial version of IFR-CS, the proposed IFR-Net network not only learns model parameters and the feature refinement operator, but also shifts the key effort from the online optimization stage to an up-front offline training phase, which significantly accelerates the reconstruction time. Compared to ADMM-Net, the CNN-based inversion blocks and feature refinement module enhance the network capability. Extensive experimental results show that the proposed IFR-Net achieves a better performance both visually and in the quantitative values compared to the traditional iterative methods including IFR-CS and several deep learning methods including ADMM-Net.

## APPENDIX

### A. Feature Refinement Operator

To obtain a feature descriptor $T$, a denoised image $u$ should be inputted first. Then, a Gaussian filter is used to blur the denoised $u$ and we can get a blurred image $u_b$. Assume $p$ and $q$ are two local image patches extracted from $u$ and $u_b$ respectively, the feature descriptor $T$ is defined as follows:

$$T(u) = 1 - |c(p,q)s(p,q)| = 1 - \left| \frac{2\sigma_{pq} + V}{\sigma_p^2 + \sigma_q^2 + V} \right| \quad (13)$$

where $c(p,q)$ calculates the reduction of contrast variation due to the degraded operation, $s(p,q)$ quantifies the structural correction between the original and degraded images. The other variables are defined as follows: $\sigma_p = (\sum_{i=1}^{N_u}(p_i - \mu_p)^2)^{1/2}$, $\sigma_q = (\sum_{i=1}^{N_u}(q_i - \mu_q)^2)^{1/2}$, and $\sigma_{pq} = \sum_{i=1}^{N_u}(p_i - \mu_p)(q_i - \mu_q)^2$, where $\mu_p = \sum_{i=1}^{N_u} p_i$, $\mu_q = \sum_{i=1}^{N_u} q_i$ and $N_u$ is the pixel number of image $u$. The constant $V$ is introduced for numerical stability and it has a great influence on the reconstruction result. In IFR-Net, the descriptor $T$ is tuned by training the constant $V$.

### B. Parameter Updating of Training Procedure

The implementation of updating the network parameters $\Theta$ in the training procedure of IFR-Net is as follows:

1) *Feature refinement module ($\mathbf{R}^{(n)}$):*

The parameter used in this module is $V^{(n)}$ in $T^{(n)}$. The gradients of loss w.r.t. the parameters can be computed as

$$\frac{\partial E}{\partial V^{(n)}} = \frac{\partial E}{\partial x_t^{(n)}} \frac{\partial x_t^{(n)}}{\partial T^{(n)}} \frac{\partial T^{(n)}}{\partial V^{(n)}},$$

where $\frac{\partial E}{\partial x_t^{(n)}} = \frac{\partial E}{\partial x^{(n+1)}} \frac{\partial x^{(n+1)}}{\partial x_t^{(n)}}$.

We also compute the gradients of the output in this layer w.r.t. its inputs as $\frac{\partial E}{\partial x_t^{(n)}} \frac{\partial x_t^{(n)}}{\partial u^{(n)}}$ and $\frac{\partial E}{\partial x_t^{(n)}} \frac{\partial x_t^{(n)}}{\partial x^{(n)}}$.

2) *Sparsity-promoting denoising module ($\mathbf{Z^{(n)}}$):*

  *United layer ($U^{(n,k)}$):*

The parameters in this layer contain $\mu_1^{(n,k)}$ and $\mu_2^{(n,k)}$. The gradients of loss w.r.t. the parameters can be computed as

$$\frac{\partial E}{\partial \mu_1^{(n,k)}} = \frac{\partial E}{\partial u^{(n,k)}} \frac{\partial u^{(n,k)}}{\partial \mu_1^{(n,k)}}, \quad \frac{\partial E}{\partial \mu_2^{(n,k)}} = \frac{\partial E}{\partial u^{(n,k)}} \frac{\partial u^{(n,k)}}{\partial \mu_2^{(n,k)}},$$

where $\frac{\partial E}{\partial u^{(n,k)}} = \frac{\partial E}{\partial u^{(n,k+1)}} \frac{\partial u^{(n,k+1)}}{\partial u^{(n,k)}} + \frac{\partial E}{\partial c_1^{(n,k)}} \frac{\partial c_1^{(n,k)}}{\partial u^{(n,k)}}$, if $k < K$;

in addition, $\frac{\partial E}{\partial u^{(n,k)}} = \frac{\partial E}{\partial x_t^{(n)}} \frac{\partial x_t^{(n)}}{\partial u^{(n,k)}}$, if $k = K$. The gradient of the layer output w.r.t. the input is computed as $\frac{\partial E}{\partial u^{(n,k)}} \frac{\partial u^{(n,k)}}{\partial u^{(n,k-1)}}$,

$\frac{\partial E}{\partial u^{(n,k)}} \frac{\partial u^{(n,k)}}{\partial x^{(n)}}, \frac{\partial E}{\partial u^{(n,k)}} \frac{\partial u^{(n,k)}}{\partial c_2^{(n,k)}}$.

  *Convolution layer ($C_1^{(n,k)}$):*

The parameters in this layer are $w_1^{(n,k)}$ and $b_1^{(n,k)}$. The gradients of loss w.r.t. the parameters can be computed as

$$\frac{\partial E}{\partial w_1^{(n,k)}} = \frac{\partial E}{\partial c_1^{(n,k)}} \frac{\partial c_1^{(n,k)}}{\partial w_1^{(n,k)}}, \quad \frac{\partial E}{\partial b_1^{(n,k)}} = \frac{\partial E}{\partial c_1^{(n,k)}} \frac{\partial c_1^{(n,k)}}{\partial b_1^{(n,k)}},$$

where $\frac{\partial E}{\partial c_1^{(n,k)}} = \frac{\partial E}{\partial h^{(n,k)}} \frac{\partial h^{(n,k)}}{\partial c_1^{(n,k)}}$.

The gradient of the layer output w.r.t. the input is computed as $\frac{\partial E}{\partial c_1^{(n,k)}} \frac{\partial c_1^{(n,k)}}{\partial u^{(n,k)}}$.

  *Nonlinear transform layer ($H^{(n,k)}$):*

The parameters of this layer are $\{q_i^{(n,k)}\}_{i=1}^{N_c}$. The gradients of loss w.r.t. the parameters can be computed as

$$\frac{\partial E}{\partial q_i^{(n,k)}} = \frac{\partial E}{\partial h^{(n,k)}} \frac{\partial h^{(n,k)}}{\partial q_i^{(n,k)}},$$

where $\frac{\partial E}{\partial h^{(n,k)}} = \frac{\partial E}{\partial c_2^{(n,k)}} \frac{\partial c_2^{(n,k)}}{\partial h^{(n,k)}}$.

The gradient of the layer output w.r.t. the input is computed as $\frac{\partial E}{\partial h^{(n,k)}} \frac{\partial h^{(n,k)}}{\partial c_1^{(n,k)}}$.

  *Convolution layer ($C_2^{(n,k)}$):*

The parameters in this layer are $w_2^{(n,k)}$ and $b_2^{(n,k)}$. The gradients of loss w.r.t. the parameters are computed as

$$\frac{\partial E}{\partial w_2^{(n,k)}} = \frac{\partial E}{\partial c_2^{(n,k)}} \frac{\partial c_2^{(n,k)}}{\partial w_2^{(n,k)}}, \quad \frac{\partial E}{\partial b_2^{(n,k)}} = \frac{\partial E}{\partial c_2^{(n,k)}} \frac{\partial c_2^{(n,k)}}{\partial b_2^{(n,k)}},$$

where $\frac{\partial E}{\partial c_2^{(n,k)}} = \frac{\partial E}{\partial u^{(n,k)}} \frac{\partial u^{(n,k)}}{\partial c_2^{(n,k)}}$.

The gradient of the layer output w.r.t. the input is computed as $\frac{\partial E}{\partial c_2^{(n,k)}} \frac{\partial c_2^{(n,k)}}{\partial h^{(n,k)}}$.

3) *Reconstruction module ($\mathbf{X^{(n)}}$):*

The parameter of this module is $\rho^{(n)}$. The gradient of the layer output w.r.t. the input can be computed as follows:

$$\frac{\partial E}{\partial \rho^{(n)}} = \frac{\partial E}{\partial x^{(n)}} \frac{\partial x^{(n)}}{\partial \rho^{(n)}},$$

where $\frac{\partial E}{\partial x^{(n)}} = \frac{\partial E}{\partial x_t^{(n)}} \frac{\partial x_t^{(n)}}{\partial x^{(n)}} + \frac{\partial E}{\partial u^{(n)}} \frac{\partial u^{(n)}}{\partial x^{(n)}}$, if $n \leq N_s$; in addition,

$$\frac{\partial E}{\partial x^{(n)}} = \frac{1}{|\psi|} \frac{x^{(n)} - x^{gt}}{\sqrt{\left\|x^{(n)} - x^{gt}\right\|_2^2} \sqrt{\left\|x^{gt}\right\|_2^2}}, \text{ if } n = N_s + 1.$$

The gradient of the layer output w.r.t. the input is computed as $\frac{\partial E}{\partial x^{(n)}} \frac{\partial x^{(n)}}{\partial x_t^{(n-1)}}$.